\def\year{2021}\relax
\documentclass[letterpaper]{article}
\usepackage{aaai21} 
\usepackage{times} 
\usepackage{helvet} 
\usepackage{courier} 
\usepackage[hyphens]{url} 
\usepackage{graphicx} 
\usepackage{graphicx} 
\usepackage{natbib}
\usepackage{caption}
\usepackage{booktabs}
\usepackage{slashbox,pict2e}
\usepackage{multirow}
\usepackage{amsmath}
\usepackage[switch]{lineno}  %
\usepackage{tabularx}
\DeclareMathOperator*{\argmax}{arg\,max}

\title{Improving Commonsense Causal Reasoning\\by Adversarial Training and Data Augmentation}

\author {
        Ieva Stali\={u}nait\.{e},
        Philip John Gorinski,
        Ignacio Iacobacci \\}

\affiliations{
    Huawei Noah's Ark Lab, London, United Kingdom\\
    \{ieva.staliunaite $|$ philip.john.gorinski $|$ ignacio.iacobacci\}@huawei.com
}

\begin{document}

\maketitle

\begin{abstract}
Determining the plausibility of causal relations between clauses is a commonsense reasoning task that requires complex inference ability. The general approach to this task is to train a large pretrained language model on a specific dataset. However, the available training data for the task is often scarce, which leads to instability of model training or reliance on the shallow features of the dataset. This paper presents a number of techniques for making models more robust in the domain of causal reasoning. Firstly, we perform adversarial training by generating perturbed inputs through synonym substitution. Secondly, based on a linguistic theory of discourse connectives, we perform data augmentation using a discourse parser for detecting causally linked clauses in large text, and a generative language model for generating distractors. Both methods boost model performance on the Choice of Plausible Alternatives (COPA) dataset, as well as on a Balanced COPA dataset, which is a modified version of the original data that has been developed to avoid superficial cues, leading to a more challenging benchmark. We show a statistically significant improvement in performance and robustness
on both datasets, even with only a small number of additionally generated data points. 
\end{abstract}

\section{Introduction}

Within the discourse discipline in linguistic research, 
causal relations are classified as \textit{contingency relations}, which in turn are a subset of coherence relations. Coherence relations are temporal, comparison, expansion and contingency relations between clauses, which can be explicitly marked with certain discourse connectives \cite{asr2013information}. Examples of the expressions of the four types of coherence relations include ``\textit{A then/before/while B}", ``\textit{A even though/but/however B}", ``\textit{A and/moreover/or B}" and ``\textit{A because/therefore/if B}", respectively. In this work we are specifically interested in the causal relations, which can be split into backward (e.g. expressed as ``\textit{A because B}") and forward (e.g. expressed as ``\textit{A therefore B}") ones, differing with regard to whether A or B refers to the cause of the event expressed by the other clause. 
It is interesting that causal connectives such as ``\textit{because}" convey a lot of information needed for inferring the causal relation between clauses. 
That is, the discourse connective ``\textit{because}" disambiguates the causal relation between the clauses more than most other connectives \cite{asr2013information}. 

From a Natural Language Processing (NLP) perspective, the task of causal inference is to determine the presence of a causal link \textit{without} the cue of a discourse connective. 
Commonsense causal reasoning is a complex task, which requires not only linguistic parsing and logical inference but also world knowledge of causal links between events. 
For example, if a model was to expertly determine which of the two alternatives is more likely to be caused by the premise below, it would have to access knowledge of the causal link between the ripening and edibility of bananas.

\begin{figure}[h!]
\noindent \textbf{Premise:} The bananas ripened. \\
\noindent \textbf{Alternative1:} We squeezed them. \\
\noindent \textbf{Alternative2:} We ate them. \\
\noindent \textbf{Question:} Effect \\
\noindent \textbf{Label:} Alternative2
\caption*{From \citet{roemmele2011choice}}
\end{figure}

The most commonly used benchmark task for evaluation of commonsense reasoning models is the the Choice of Plausible Alternatives \cite[COPA]{roemmele2011choice}. 
In this task the goal is to determine which of the two given alternatives is the \textit{true} choice, i.e. which alternative is causally linked to a given premise, and which one is a \textit{distractor}. This is an easier task than determining the precise coherence relation between two clauses. 
The task is relatively easy for humans, as shown by the nearly perfect annotator agreement for this task (Cohen's kappa = 0.965). However, the COPA dataset is very small (1000 items in total), and high model performance on it is not stable and often relies on biases in the data \cite{kavumba2019choosing}.
Consequently, \citet{kavumba2019choosing} 
introduce
a Balanced COPA dataset by manually adjusting items from COPA to remove the superficial features, such as patterns of determiner use, which are generally shown to be exploited by models for solving the task. 
Data points are generated by mirroring each example in a way that it captures a similar relation, yet with the biasing feature appearing in the alternative which does not manifest that feature in the original data. That way the biasing feature appears in the \textit{true} choice in one case and in the \textit{distractor} in the other case. 
For example, \citet{kavumba2019choosing} show that models rely on indefinite determiners like ``\textit{a}" when classifying instances, as their distribution is not uniform between the classes. A mirrored example is shown here: 

\begin{figure}[h!]
\noindent \textbf{Original premise:} The woman hummed to herself.\\
\noindent \textbf{Original alternative1 (true):} She was in \textbf{a} good mood.\\
\noindent \textbf{Original alternative2 (distractor):} She was nervous.\\
\noindent \textbf{Mirrored premise:}  The woman trembled.\\
\noindent \textbf{Mirrored alternative1 (distractor):}  She was in \textbf{a} good mood.\\
\noindent \textbf{Mirrored alternative2 (true):}  She was nervous.
\caption*{From \citet{kavumba2019choosing}}
\end{figure}

This leads to a dataset which is double the size of the original COPA dev set and more challenging to solve than the original COPA. 

As briefly discussed by \citet{gordon2011commonsense}, the research on causal inference tasks can be split into approaches that focus on the \textit{depth} or \textit{breadth} of the solution. Deep approaches aim to gain more information about a particular input,
for example by using knowledge graphs to learn more about the entities and events mentioned in the input or applying formal logic to deterministically find the relation that is sought for, in order to classify that input correctly \cite{furbach2015tackling, furbach2016commonsense, blass2017analogical, siebert2019commonsense, goodwin2019bridging}. 
On the other hand, the approaches of the broad type attempt to cover a wider range of instances in their method, for example by determining the types of syntactic or semantic features that are generally used in expressing a particular relation \cite{gordon2011commonsense, goodwin2012utdhlt, jabeen2014using, rahimtoroghi2017learning, tamborrino2020pre, iter2020pretraining}.
This paper attempts to tackle the causal reasoning task with two approaches, one of the \textit{deep} and one of the \textit{broad} type.
That is, the first approach is adversarial training - perturbing the original inputs to produce similar but more difficult examples, which can be seen as data points in the surrounding area of the original ones. The aim of covering the area around the original datapoints is to provide the model with more semantic information about the inputs. 
The second approach is
data augmentation by means of generating completely unseen examples, which leads to new data points further away from the original ones. The aim of covering a wider range of examples is to find general patterns of causally linked clauses. 

In both methods we study to what extent model performance on the task of causal inference can be improved by augmenting the training data with linguistic information. To this end, we rely on findings of psycholinguistic research, discourse annotations, lexical semantics and language models for tackling the COPA task, using a RoBERTa model \cite{liu2019roberta} as our baseline. The first approach is an application of an adversarial example generation through perturbations of the original COPA data by substituting words with their equivalent terms given the context using a semantic network, following \citet{zang2020word}. The second approach consists of gathering causally linked clauses from the web with the help of a Penn Discourse TreeBank 
\cite[PDTB]{prasad2008penn}
parser \cite{lin2014pdtb} and generating the distractor alternatives by applying a technique based on discourse connectives and their denotations with the help of GPT-2 \cite{radford2019language}. Both approaches lead to varying improvements on the performance of the RoBERTa model on both the COPA and balanced COPA datasets, exhibited by higher average accuracy scores and smaller standard deviation ranges. 
The three main contributions in this paper 
can be summarized as: 

\bigskip
\noindent \textbf{1.} A novel application of linguistic knowledge to the task of causal reasoning; \\
\noindent \textbf{2.} A productive augmentation method for the task of choosing plausible causal alternatives; \\
\noindent \textbf{3.} A significant improvement on the performance and robustness of a RoBERTa model on the COPA dataset. \\

\section{Related Work}

A wide variety of approaches have been used for tackling the task of COPA. The previously used \textit{deep} approaches include formal linguistic methods such as the use of theorem proving on text converted to a logical form, and structured knowledge methods such as the use of knowledge graphs. 
The \textit{broad} approaches include more heuristic methods such as assuming the presence of causal relations between co-located sentences in narrative text, and Neural Network approaches such as adjusting loss functions or pre-training language models with an unconventional target. 

The formal linguistic representation approaches
\cite{furbach2015tackling, furbach2016commonsense, blass2017analogical, siebert2019commonsense} vary with regard to their focus - from tackling the problem of chaining multiple logical inferences by analogy, to merging the formal representation approach with a Neural Network model. While the theory proving approach guarantees the correct outcome given the correct representation, the representations are not easy to produce, especially for colloquial use of language. 

For incorporating background knowledge, some previous research has adopted structured knowledge approaches such as knowledge graphs 
\cite{goodwin2019bridging} 
and semantic networks 
\cite{mtarji2019commonsense}. 
Knowledge graphs about entities as well as causal links between event words provide information that is relevant to the type of reasoning required for commonsense causal reasoning tasks, which is assumed to be emitted in raw text based on pragmatic rules of language. 
Hence, as expected, the use of external world knowledge improves the performance of the prevailing neural network systems. 

In addition, 
external linguistic knowledge sources have also been used for training models, which can be combined with other systems to inject linguistic information into the causal reasoning systems as well. 
For example, \citet{goodwin2012utdhlt} use syntactic dependency trees as well as annotations of temporal relations between events. In the above mentioned approaches the authors rely on assumptions such as that an event \textit{A} is not likely causally linked to event \textit{B} if words in \textit{A} are not conceptually linked to words in the context of \textit{B}, that
causal relations are exhibited in certain syntactic patterns in text, and that causal relations tend to fall into a temporally bound pattern. 
Finally, some research uses a discourse parser trained on manually annotated causally linked pairs of clauses and sentence similarity metrics to determine the more likely causal links \cite{gordon2011commonsense}. This approach is based on the expectation that similar sentences will stand in similar causal relations. 

The latter approach differs from the aforementioned ones in that it does not annotate the words or sentences in the target clauses, but relies on other input through similarity. 
Similarly, a slightly different set of assumptions allows one to use only raw text or more coarse annotations in training causal reasoning systems with some fruitful outcomes as well. For example, a few papers present work based on the idea that narrative text is composed of sequences of causally linked sentences. \citet{gordon2011commonsense}, in addition to the previously discussed method, also train a model on sentences from personal blogs that appear in close proximity. The hypothesis in this method is that there is some link between the proximity of sentences in personal stories as people tend to tell them in causally linked sequences.
In a very similar vein, \citet{rahimtoroghi2017learning} propose learning which events are contingent on others by expecting that the sentences describing the \textit{effects} appear \textit{after} sentences describing the \textit{causes} in a narrative. They show that they are able to learn some new relations that were not present in structured resources at the time. 
Similarly, \citet{jabeen2014using} show that the order in which two events appear is relevant to finding contingency relations, as asymmetrical labels of events lead to better causal inference results than symmetrical ones. 

Finally, large pre-trained Neural Network models are very prevalent in NLP tasks including causal reasoning, with the current state-of-the-art result on COPA belonging to the T-5 model finetuned for the downstream task \cite{raffel2019exploring}. 
Beside the use of out-of-the-box models, some research has proposed a different pre-training technique that is specifically aiming to improve discourse tasks. More specifically, \citet{iter2020pretraining} use the distance between sentences as an additional training task, under the same assumption as the research discussed in the previous paragraph - namely the expectation of causal relations between subsequent sentences. They show an improvement on discourse-related tasks such as COPA among others, when a BERT model \cite{devlin-etal-2019-bert} is trained with this additional objective. 
In addition, \citet{shwartz2020unsupervised} use a self-talk method in which they use pre-trained generative language models to produce questions and answers about entities mentioned in the target sentences in a causal reasoning task. This provides the background knowledge in a novel way, as an alternative to the use of knowledge graphs mentioned at the beginning of this section.
In contrast, \citet{tamborrino2020pre} rephrase the causal reasoning task into an ``\textit{A because B}" format and calculate the likelihood of each token in the input by masking one at a time, leading to much higher performance when compared to the use of the same model for fine-tuning on the target task.

In this work we combine the ideas of basing the assumptions on the results of experimental linguistics as well as employing a generative language model in an unconventional way, in order to benefit both from linguistic research and the data-driven approaches.

\section{Causal Relation Classification Model}
We give a brief description of the model architecture we employ for all experiment settings in the later sections. We treat COPA as a \textit{sequence classification problem}, in which we want to predict for an input sentence $s$, consisting of a premise $p$ and choice $c$, a label $l\in\{0,1\}$, indicating whether $p$ and $c$ are connected by a \textit{causal} relation. While the original COPA data contains both \textit{cause} and \textit{effect} relations, we make use of the same property observed by \citet{li-etal-2019-learning}, namely that the relation between any premise $p$ and continuation $c$ is \textit{reversible}. For example, for the premise-choice pair \textit{``The woman's date wanted to look like a gentleman.''}, \textit{``He opened the door for her.''} that is connected in an \textit{effect} relation (\textit{``The woman's date wanted to look like a gentleman, \textbf{therefore} he opened the door for her.''}), we can treat the reverse relation as \textit{causal}: \textit{``He opened the door for her, \textbf{because} the woman's date wanted to look like a gentleman.''}\footnote{Arguably, this sentence is grammatically marked, as the pronoun appears before its antecedent, however we consider this drawback minor in comparison to the gain of unifying all examples into the same relation type.}, by switching the order of premise and choice sentences. This simplifies our classification problem, as the model needs only learn to predict a single relation.

\begin{figure}[t]
    \centering
    \includegraphics[width=.95\columnwidth]{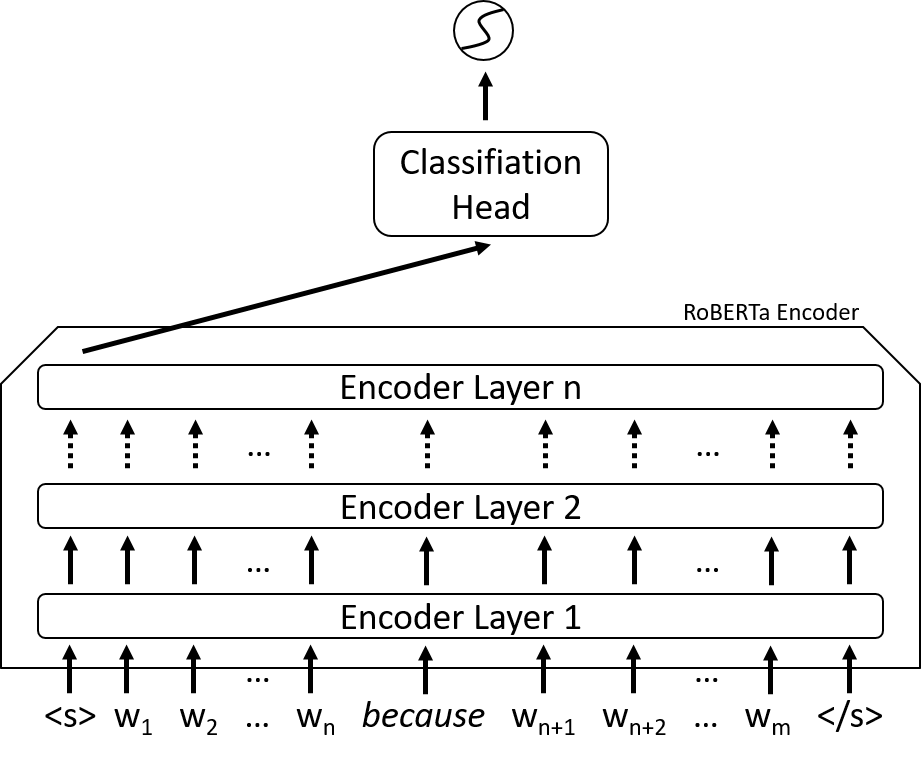}
    \caption{Our classification architecture used to predict COPA relations. A converted premise-choice sentence is encoded with a pre-trained RoBERTa model, and classified using a simple feed-forward network.}
    \label{fig:classifier}
\end{figure}

For our classification model, we use the implementation of RoBERTa \cite{liu2019roberta} provided by Huggingface\footnote{\url{https://huggingface.co/}} to encode a given input sentence, with a simple linear classifier that takes the final encoding of RoBERTa's \texttt{<s>} token as input. Figure~\ref{fig:classifier} shows a depiction of our model architecture. For all our experiments, we use \textit{RoBERTa\_large} as the encoder model, and a single-layer linear classifier with $1024$-dimensional input and 2 output dimensions, with subsequent softmax.

During model training, we use the cross-entropy loss as optimization objective to predict $P(label|conv(p,c,r))$, where $p$ and $c$ are the original premise and a single choice, $r$ is the original relation between $p$ and $c$, $label\in\{0,1\}$, and $conv(p,c,r)$ is a function that converts the premise and choice into an input suitable for RoBERTa, similar to \citet{li-etal-2019-learning}:
\[
conv(p,c,r)=
 \begin{cases}
 \text{\textless s\textgreater}\ w^p_1\ldots w^p_n\ \text{because}\ w^c_1 \ldots w^c_n\ \text{\textless /s\textgreater} & \\
 \text{\hspace*{2.2em}, if r}=\text{cause} &\\
 \text{\textless s\textgreater}\ w^c_1\ldots w^c_n\ \text{because}\ w^p_1 \ldots w^p_n\ \text{\textless /s\textgreater} & \\
 \text{\hspace*{2.2em}, if r}=\text{effect} &
 \end{cases}
\]

At inference time, we take as correct choice the one which, under the trained model, satisfies: \[c^*=\argmax_{c\in\{\text{choice1, choice2}\}}P(1|conv(p,c,r))\]

Similarly to \citet{kavumba2019choosing}, we use the learning rates of $1e-6$, $2e-6$ and $3e-6$, 20 different seeds per learning rate, weight decay of 0.01, and a batch size of 32. We train for a maximum of 50 epochs, stopping early when performance on the development set ceases to improve. We average the model performance on the development data over the results of the 20 seeds per learning rate, remove the bottom and top two outliers, and evaluate those models trained with the best-performing learning rate on the test set. 

Wherever we train models on original or augmented datasets, we assume this very model architecture and training scheme. In addition, in result tables and discussions, we will refer to the original COPA training data as \textit{Base}, for brevity.

\section{Augmentation with Adversarial Examples}

Given that the COPA dataset is very small, there is a need for methods that make models trained on it perform better with regard to both highest possible accuracy and robustness. 
This section presents work on using adversarial training by \textit{perturbing} the original training set input sentences slightly, and using the sentences that make the trained model fail as additional data points. 

Following the framework of \citet{zang2020word} we adapt their approach to the task of COPA with various modifications. 
\citet{zang2020word} work on the natural language inference and sentiment analysis tasks using a BERT-based model, adversarially attacking the inputs, and show that using the perturbed inputs as additional training data makes the BERT model more robust. 
The input perturbation consists of substituting some content words in the input with other words that share the same basic semantic units with the original word. The authors filter the potential substitution words by only using those that can appear as the same part of speech as the original word using HowNet \cite{qi2019openhownet} as a resource of word substitutions. 
Since computing all the combinations is intractable, \citet{zang2020word} then use a Particle Swarm Optimization \cite[PSO]{kennedy1995particle} algorithm to find the cases in which the model correctly classifies the original instance, but fails on the perturbed one.

In adapting this approach to the task at hand, we perform the following adjustments in our implementation: 

\noindent \textbf{Knowledge Base:} we replace HowNet with WordNet \cite{fellbaum2012wordnet} due to the resource size and quality.

\noindent \textbf{Adversarial Attacks:} we replace Particle Swarm with Ant Colony Optimization \cite[ACO]{dorigo1992optimization} for choosing the best perturbation, with the aim of finding the most optimal adversarial examples. ACO is an algorithm inspired by biology -- the way ants find their way around their environment -- and belongs to the category of population-based search algorithms, which also includes PSO. It is a type of a metaheuristics technique \cite{talbi2009metaheuristics} with a focus on exploration space rather than optimization. ACO is able to find the \textit{best path} between two points in a graph, which in our case are the first and final words in a sentence, and the paths are all the potential lexicalizations of each token;

\noindent \textbf{Sentence Length:} shorter sentences are included in the perturbations due to the short average length of COPA sentences (see also Table~\ref{tab:copa_stats}); 

\noindent \textbf{Semantic Substitution:} potential substitution words are only selected if the substituting word has the same sense as the original one, in order to ensure that the words are only substituted with their synonyms which are relevant given the context, so as not to change the meaning of the sentence. To this end we perform Word Sense Disambiguation using SupWSD\footnote{\url{https://supwsd.net/supwsd/}} introduced by \citet{papandrea2017supwsd}. 

\noindent \textbf{Pretrained Model:} We replace the BERT model used by \citet{zang2020word} with RoBERTa as the base model, as described previously, due to it achieving higher results than BERT on a variety of Natural Language Processing tasks.

With these modifications, applying our adversarial attack method to the COPA training set leads to an 11.82\% success rate in failing a previously trained RoBERTa-based COPA classification model where it originally classifies the inputs correctly. These successful attacks result in a set of 76 additional training items which we merge into the original training data, including changes in either the premise or one of the alternatives such as the following: 

\bigskip
\noindent \textbf{Original:} The man craved a \textit{cigarette}. He was addicted to nicotine. \\
\noindent \textbf{Perturbed:} The man craved a \textit{butt}. He was addicted to nicotine.

\bigskip
\noindent While most generated perturbations seem to be of a high quality, there are occasional examples of either word sense disambiguation or subsequent replacement failing, for example in:

\bigskip
\noindent \textbf{Original:} The dog emitted a foul smell. The \textit{skunk} sprayed the dog.\\
\noindent \textbf{Perturbed:} The dog emitted a foul smell. The \textit{lowlife} sprayed the dog. 

\bigskip
\noindent We leave the task of improving the automatic disambiguation to future work, but note that such blatantly wrong perturbations are a small minority of cases.

\subsection{Adversarial Attack Evaluation}

To evaluate the impact that our adversarially generated data has on model performance, we train classification models as described above on the base data set as well as on the data augmented with adversarial examples. Table~\ref{tab:baseresults} summarizes the results of model performance.
\begin{table}[t]
    \centering
    \begin{tabular}{l|c|c|c|c|c}
        \toprule
        & \#Train & Min & Max & Mean & Std \\
        \midrule
        \multicolumn{6}{c}{COPA Dev}\\
        \midrule
        Base & 400 & 82.80 &88.80  &\multicolumn{1}{l|}{86.21}  &1.84\\
        +Adversarial & 476 & \textbf{85.80} & \textbf{89.20} & \textbf{87.35}* & \textbf{1.10}\\
        \midrule
        \multicolumn{6}{c}{COPA Test}\\
        \midrule
        Base & 400 &79.00  &91.00  &\multicolumn{1}{l|}{85.69}  &4.11 \\
        +Adversarial & 476 & \textbf{85.00} & \textbf{92.00} & \textbf{88.19}* & \textbf{2.40}\\
        \midrule
        \multicolumn{6}{c}{Balanced COPA Dev}\\
        \midrule
        Base & 400 & 76.50 &86.00  &\multicolumn{1}{l|}{81.72}  &2.64 \\
        +Adversarial & 476 & \textbf{79.50} & 86.00 & \textbf{82.84}* & \textbf{1.49}\\
        \bottomrule
    \end{tabular}
    \caption{The results of the adversarially trained model compared to the baseline. * indicates a significant improvement over the Base model (p $<$ 0.001).}
    \label{tab:baseresults}
\end{table}

The adversarially enhanced models outperform the models trained on the original data alone in terms of both average performance and standard deviation. 
Hence, adversarial training leads to both higher performance and higher robustness of the model with as little as 76 additional data points. The improvements are statistically significant (p $<$ 0.001) according to adapted approximate randomization \cite{noreen1989computer} provided by \citet{roemmele2011choice}.

\section{Augmentation by Causal Sentence Extraction and Distractor Generation}

While the COPA dataset is very well curated, it only contains $1,000$ examples in total, providing an extremely low-resource setting, and generating more such high-quality data by hand would be quite slow and laborious. In this section we propose a method for automatically finding new data for augmentation from large free-form text. 
This augmentation involves three important steps, described in detail in the remainder of this section.

\subsection{Data Augmentation Approach}

\noindent\textbf{Linguistically-motivated Filtering Strategy.} As mentioned above, the original COPA dataset contains examples with \textit{cause} and \textit{effect} relations. In order to find relevant new sentences in large data, we require a filtering strategy that effectively extracts such \textit{cause-effect} relations. To this end, we draw from linguistic theories to motivate strict criteria for filtering.

As with many discourse relations, \textit{cause} and \textit{effect} can be expressed either \textit{implicitly} - through the content of the sentences, or \textit{explicitly} - through the use of discourse connectives. 
In this work we rely on the explicit expression of such relations for finding causally linked clauses in raw web text. 
We define a total of 8 causal connectives\footnote{as a result, because, if, since, so, therefore, thus, when}, including forward- and backward-projecting instances.

While the defined connectives will yield generally appropriate sentences, the original COPA dataset imposes a number of additional restrictions on our desired augmented data. In particular, COPA contains premise-choice pairs that on average form relatively short sentences, and are balanced with regard to the lengths of the premise and the choices. Some length statistics of the original dataset are shown in Table~\ref{tab:copa_stats}.
\begin{table}[t]
    \centering
    \begin{tabular}{l|r|r|r|r|r}
    \toprule
         & Min & Max & Mean & Median & Std\\
         \midrule
         Premise & 2 & 13 & 6.2 & 6 & 1.8\\
         Choice 1 & 2 & 11 & 5.1 & 5 & 1.6\\
         Choice 2 & 2 & 11 & 5.1 & 5 & 1.6\\
         Total Length & 5 & 22 & 11.2 & 11 & 2.4 \\
         Premise/Total & 22.2 & 82.8 & 54.7 & 54.5 & 0.1\\
    \bottomrule
    \end{tabular}
    \caption{COPA dataset word counts. ``Total Length'' is averaged over all premise-choice pairs. The last row shows the ratios of premise to total length.}
    \label{tab:copa_stats}
\end{table}
In accordance with these properties, we refine our filtering strategy to only consider sentences between 5 and 22 words long, with a connective in the center $\pm$2 words.
Finally, we observe that neither premises nor choices in COPA contain a (potential) \textit{cause-effect} connective themselves. We therefore adapt our filter to reject sentences that contain more than one of our defined discourse connectives.

Furthermore, we also filter out vaguely phrased sentences such as ``\textit{My store doesn't do that because that method can be unreliable.}". The latter sentence does not contain much information, as it refers to actions and entities outside of the sentence, and only vaguely refers to the main events of the two clauses with verbs like ``\textit{do}" and ``\textit{be}". In order to filter such sentences out, we only consider sentences which contain ``\textit{implicitly causal verbs}". Implicit causality refers to a characteristic that certain verbs exhibit, namely that part of their meaning implies a causal directionality \cite{garvey1974implicit}. For instance, the verb ``\textit{apologize}" is implicitly causal as it pertains to preferences to attribute the cause of the apology to the subject of the verb, meaning that when processing a sentence containing this verb, people tend to assume that the cause of the apology is an action performed by the person apologizing. In this work, sentences are filtered out if they do not contain any implicitly causal verb as experimentally determined by \citet{ferstl2011implicit}.

\noindent\textbf{Large Free-form Text Resources.} Since our criteria for candidate augmentation sentences are quite strict, we expect only a small amount of any available text resource to meet them. Additionally, sentences in COPA are quite varied in terms of their (perceived) source, ranging from sentences likely encountered in childrens' books to ones most likely sourced from newspaper text. Using traditional standard single-domain datasets for the augmentation task thus seems sub-optimal. We therefore opt to use the recently published OpenWebText corpus \cite{Gokaslan2019OpenWeb}, itself derived from a non-open dataset introduced in \citet{radford2019language}. OpenWebText contains $\sim$40GB of text from over 8 million documents, spanning a plethora of resources and domains. These properties make it an ideal resource for our strict COPA sentence filtering approach.

\noindent\textbf{Data Augmentation Tool Chain.} With a text resource and filtering strategy in place, we set up our tool chain for data augmentation as follows. First, we extract from OpenWebText a potentially large number of sentences that conform to the requirements set out before. To ensure the validity of the extracted sentences and of them encoding a \textit{cause/effect} relationship, we then analyze each sentence using the Penn Discourse Treebank parser by \citet{lin2014pdtb}, and reject all sentences that either fail to parse, or do not parse as having the desired relation. After all filtering steps and PDTB parsing, we have obtained valid sentences constituting \emph{positive} examples of a premise and a single choice. 

To generate the distractor choices, we devise three strategies. For a given \emph{premise-choice pair} we either (i) randomly choose the second choice from OpenWebText; (ii) choose it from OpenWebText if it shares at least one content word with the premise; (iii) use GPT-2 \cite{radford2019language} to automatically generate a distractor.
The (i) \textit{random} selection of the distractor assumes a low probability of accidentally choosing a sentence that happens to be causally linked to our premise. While being the simplest heuristic, this method leads to examples which are potentially relatively easy to solve, as features such as sentence similarity could be expected to discriminate between the true cause and a random sentence from the web. 
The (ii) \textit{overlap} selection method assumes that alternatives which overlap in their content words with the premise would be semantically more similar and therefore potentially more difficult for models to solve. 
Finally, the (iii) \textit{GPT-2} method assumes the distinct denotations of certain discourse connectives, as discussed below. 

For generating alternatives automatically with GPT-2, we make use of the same cause-effect reversibility discussed previously, re-writing all filtered augmentation data to encode backward causal relations. Then, the first part of the resulting sentence is selected up to but not including ``\textit{because}" and the conjunction connective is added as a new sentence start ``\textit{. And}" to condition the language model. 
The second half of the sentence is subsequently generated by GPT-2, subject to the aforementioned length constraints. We finally take the resulting continuation as the distractor choice of the augmented data point.
The conjunction connective is chosen here because whenever it denotes temporal relations, they tend to be forward relations. That is, sentences of the form ``\textit{A and B}'', where \textit{A} and \textit{B} are clauses, usually refer to chronologically ordered events, where \textit{A} happens before \textit{B}. In such cases, \textit{B} can be an effect of \textit{A}, however it is unlikely to be the cause of \textit{A}. Such a forward relation can be employed here due to the fact that all the causal relations are converted into backward ones, which leads to the true and distractor alternatives denoting vastly different types of discourse relations - by type (causal vs. expansion) as well as direction (backward vs. forward). 
Thus, with this linguistically informed technique, we generate distractors which are semantically related to the premise based on the fact that they are generated as conjuncts to the premise. Between our proposed generative augmentation approaches, we hypothesize this to be the most difficult type of example for a causal inference model to distinguish, and thus conjecture it to yield the best results when used for data augmentation, compared to the other two methods.

\subsection{Experiments with Augmented Data}
To evaluate the impact of our extractive augmentation approach on COPA performance, we carry out comparative experiments between setups using the different methods of generating new data.

We train our general classification architecture on data derived from the augmentation strategies described above. For this, we first generate $400$ new training instances per generation method (random, overlap, and GPT-2), and train models on the original COPA data, individual augmented datasets, and the merged sets of augmented and original data.
Table~\ref{tab:extract_results} shows results when training on the individual datasets, as well as on combined original and augmented data.

\begin{table}[ht]
    \centering
    \begin{tabular}{l|c|c|c|c|c}
        \toprule
        & \#Train & Min & Max & Mean & Std\\
        \midrule
        \multicolumn{6}{c}{COPA Dev}\\
        \midrule
        Base & 400 & 82.80 &88.80  &\multicolumn{1}{l|}{86.21}  &1.84 \\
        Random & 400 & 60.40 & 72.40 & \multicolumn{1}{l|}{68.51} & 3.18 \\
        \hspace{1em}+Base & 800 & 84.80 & 88.80 & \multicolumn{1}{l|}{86.73} & 1.41\\
        Overlap & 400 &60.40  &70.40  &\multicolumn{1}{l|}{66.01}  &3.49 \\
        \hspace{1em}+Base & 800 & 85.40 & 88.60 & \multicolumn{1}{l|}{86.74} & \textbf{1.07}\\
        GPT-2 & 400 & 64.60 & 75.00 & \multicolumn{1}{l|}{71.25} & 2.84 \\
        \hspace{1em}+Base & 800 & 83.60 & \textbf{89.20} & 86.93* & 1.75\\
        All & 1200 & 74.44 & 79.80 & \multicolumn{1}{l|}{76.85} & 1.71 \\
        \hspace{1em}+Base & 1600 & \textbf{85.80} & \textbf{89.20} & \textbf{87.75}* & 1.16 \\
        \midrule
        \multicolumn{6}{c}{COPA Test}\\
        \midrule
        Base & 400 &79.00  & 91.00 & \multicolumn{1}{l|}{85.69}  &4.11 \\
        Random & 400 & 57.00 & 71.00 & \multicolumn{1}{l|}{63.63} & 4.47 \\
        \hspace{1em}+Base & 800 & 78.00 & 93.00 & \multicolumn{1}{l|}{86.33} & 3.94\\
        Overlap & 400 &57.00 &73.00 & \multicolumn{1}{l|}{65.19} &4.65 \\
        \hspace{1em}+Base & 800 & \textbf{86.00} & 90.00 & \multicolumn{1}{l|}{88.13} & \textbf{1.67}\\
        GPT-2 & 400 & 62.00 & 77.00 & \multicolumn{1}{l|}{71.31} & 3.65 \\
        \hspace{1em}+Base & 800 & \textbf{86.00} & \textbf{94.00} & \textbf{90.24}* & 2.28\\
        All & 1200 & 70.00 & 83.00 & \multicolumn{1}{l|}{77.13} & 3.65 \\
        \hspace{1em}+Base & 1600 & 85.00 & 92.00 & 89.44* & 1.71\\
        \midrule
        \multicolumn{6}{c}{Balanced COPA Dev}\\
        \midrule
        Base & 400 & 76.50 &86.00  & \multicolumn{1}{l|}{81.72}  &2.64 \\
        Random & 400 & 60.00 & 71.50 & \multicolumn{1}{l|}{63.31} & 3.31 \\
        \hspace{1em}+Base & 800 & 75.50 & \textbf{86.50} & \multicolumn{1}{l|}{82.80} & 2.78\\
        Overlap & 400 & 57.00& 71.50 & \multicolumn{1}{l|}{64.16} & 3.51 \\
        \hspace{1em}+Base & 800 & \textbf{80.00} & 85.00 & \multicolumn{1}{l|}{82.03} & \textbf{1.50}\\
        GPT-2 & 400 & 66.00 & 72.50 & \multicolumn{1}{l|}{69.31} & 2.15\\
        \hspace{1em}+Base & 800 & 79.50 & 86.00 & \textbf{83.94}* & 1.91\\
        All & 1200 & 68.50 & 78.50 & \multicolumn{1}{l|}{73.44} & 2.66 \\
        \hspace{1em}+Base & 1600 & \textbf{80.00} & \textbf{86.50} & 83.78* & 1.74\\
        \bottomrule
    \end{tabular}
    \caption{The results of the model trained on data augmented through causal sentence extraction and distractor generation, compared to the baseline of trained only on original COPA data. * indicates a significant improvement over the Base model (p $<$ 0.001).}
    \label{tab:extract_results}
\end{table}

\begin{figure}[t]
  \centering
    \includegraphics[width=.95\columnwidth]{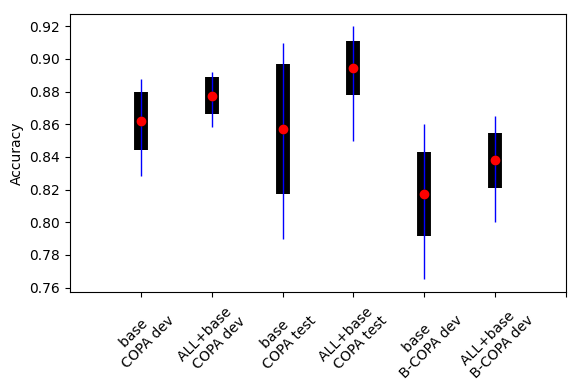}
\caption{Results of the baseline model (trained on the original COPA training set) and the best performing model (trained on GPT-2 generated distractors together with the random and overlap methods in addition to the COPA training data), evaluated on the original COPA dev set, test set and the Balanced COPA dev set.}
\label{fig:flow}
\end{figure}

Across the board, we observe that training with augmented data in addition to the original COPA data is able to boost model performance over training on just the Base data alone. It is also worth noting that all augmentation methods \textit{on their own} seem to be able to provide at least a decent learning signal, with GPT-2 and all combined data even reaching scores up to and above the $70$s on the dev, test, and balanced sets. When training with all augmentation data, models on average reach as high as $77\%$ on the dev and test sets, and around $73\%$ on balanced dev.

When adding the original COPA to the augmented data sets, we consistently achieve higher model performance. A single one of our models trained on COPA alone achieves $91\%$ on the test set, however, this comes at the price of a severe standard deviation of over $4$ points -- a deviation only ``topped'' by models trained on the random or overlap based data \textit{alone}. The single best model according to the test set is based on GPT-2-generated data in addition to the COPA training data, reaching $94\%$ accuracy. When training on all data combined, the mean results on the COPA test set are only slightly lower. On the balanced COPA dev set, again, \textit{GPT-2+Base} achieves the highest mean model performance, while the highest maximum is shared between the model based on all available data and the Random+Base model, while the highest minimum score is achieved by the model trained on all data. 
This goes to show that seemingly, our augmentation strategies successfully help overcome the base models' reliance on -- and exploitation of -- the superficial cues described previously, boosting performance on data that features no such cues.

One important improvement, also reflected in Figure~\ref{fig:flow}, is that training with augmented data in addition to COPA greatly stabilizes model training, and consistently improves the models' standard deviation in almost all cases, especially on the test and balanced dev sets. Training on just the original COPA data in fact yields very unstable performance on those data sets, with a deviation of over $4\%$ with individual evaluations ranging from $79\%$ -- $91\%$ on test, and $76.5\%$ -- $86.00\%$ on balanced dev. This renders actual model performance rather luck-based, relying heavily on randomness. With the use of augmented data, we are able to greatly stabilize the models. This leads to just over $1/3$ of COPA's original standard deviation on the test set, and just under one point decrease on the balanced dev set, when training with all available data. On the balanced set, our overlap-based augmentation almost halves the original standard deviation.

While all the augmented datasets (\textit{Random+Base}, \textit{Overlap+Base} and \textit{GPT-2+Base}) lead to higher model performance and lower standard deviation scores in most settings, the \textit{GPT-2+Base} augmentation outperforms the other two methods, and is the dataset that consistently yields significant improvements, on par with the models trained on all available data. 

Overall, our experiments suggest that our augmentation strategies are able to generate data that contains high-quality ``COPA-like'' examples, useful for improving model performance over training with the hand-crafted and curated dataset alone.

\section{Conclusions}

In this work we have explored novel applications of linguistic knowledge to the task of detecting plausible causal relations between clauses. We have introduced a set of data augmentation techniques that can be applied by adversarially attacking existing models to find weak spots, or by using generative models to produce entirely new data. Our methods are able to generate high-quality data to augment the originally available training set of COPA. Experiments with data derived from our adversarial attack and various augmentation strategies show that our methods can help make model training more robust while also improving performance.

In the future, we would like to adapt our methods to further (low resource) data sets, to explore the general viability of our adversarial and augmentation strategies for improving model performance.

\bibliography{Bibliography.bib}

\end{document}